\pdfoutput=1

\documentclass[11pt]{article}

\usepackage{color}
\usepackage{array}
\usepackage{CJKutf8}
\usepackage{booktabs}
\usepackage{stfloats}
\usepackage{amsmath}
\usepackage{graphicx}
\usepackage{setspace}
\usepackage{multirow}
\usepackage{subfigure}
\usepackage{amssymb}
\usepackage{listings}
\usepackage{enumitem}
\usepackage{tablefootnote}
\usepackage{lipsum}
\usepackage[bottom]{footmisc}

\definecolor{darkred}{RGB}{222,72,43}
\definecolor{lightred}{RGB}{255,236,232}
\definecolor{darkgreen}{RGB}{0,109,79}
\definecolor{lightgreen}{RGB}{224,251,241}
\definecolor{textgreen}{RGB}{0,128,0}
\definecolor{textred}{RGB}{255,0,0}

\usepackage{acl}
\usepackage{times}
\usepackage{latexsym}

\usepackage[T1]{fontenc}

\usepackage[utf8]{inputenc}

\usepackage{microtype}

\title{DuQM: A Chinese Dataset of Linguistically Perturbed Natural Questions for Evaluating the Robustness of Question Matching Models}

\author{\textbf{
Hongyu Zhu\textsuperscript{1}\thanks{\llap{}\:\:\:Equal contribution.\:\:The work was done when Hongyu Zhu was doing internship at Baidu.} ,
Yan Chen\textsuperscript{2}\footnotemark[1], 
Jing Yan\textsuperscript{2}\footnotemark[1], 
Jing Liu\textsuperscript{2},
Yu Hong\textsuperscript{1}, 
Ying Chen\textsuperscript{2},
} \\
\textbf{
Hua Wu\textsuperscript{2} ,
Haifeng Wang\textsuperscript{2}
}\\
    \textsuperscript{1}School of Computer Science and Technology, Soochow University, China \\  
    \textsuperscript{2}Baidu Inc., Beijing, China\\
    \{hines.zhu, tianxianer\}@gmail.com\\
    \{chenyan22, yanjing09, liujing46, chenying04, wu\_hua, wanghaifeng\}@baidu.com
}

\begin{document}
\maketitle

\begin{CJK*}{UTF8}{gbsn}

\begin{abstract}
In this paper, we focus on the robustness evaluation of Chinese Question Matching (QM) models. 
Most of the previous work on analyzing robustness issue focus on just one or a few types of artificial adversarial examples. Instead, we argue that it is necessary to formulate a comprehensive evaluation about the linguistic capabilities of QM models on natural texts. 
For this purpose, we create a Chinese dataset namely DuQM which contains natural questions with linguistic perturbations to evaluate the robustness of QM models. 
DuQM contains 3 categories and 13 subcategories with 32 linguistic perturbations. The extensive experiments demonstrate that DuQM has a better ability to distinguish different models. Importantly, the detailed breakdown of evaluation by linguistic phenomenon in DuQM helps us easily diagnose the strength and weakness of different models. 
Additionally, our experiment results show that the effect of artificial adversarial examples does not work on the natural texts.
The dataset and baseline codes will be publicly available in the open source community.
\end{abstract}

\section{Introduction}

The task of \textit{Question Matching (QM)} aims to identify the question pairs that have the same meaning, and it has been widely used in many applications, e.g., community question answering and intelligent customer services, etc. 
Though neural QM models have
shown compelling performance on various datasets, including Quora Question Pairs (QQP)~\cite{iyer2017first},  LCQMC~\cite{liu2018lcqmc}, BQ~\cite{chen2018bq} and AFQMC\footnote{It is from Ant Technology Exploration Conference (ATEC) Developer competition, which is no longer available.}, 
neural models are often not robust to adversarial examples, 
which means that the neural models predict unexpected outputs given just a small perturbations on the inputs.
As the example 1 in Tab.~\ref{tab:statistics-exp} shows, a model might not distinguish the minor difference (\textit{"面~noodles"}) between the two sentences, and thus predicts the two questions semantically equivalent.

Recently, it attracts a lot of attentions from the research community to deal with the robustness issues of neural models on various NLP tasks, such as question matching, natural language inference and machine reading comprehension. 
Early works examine the robustness of neural models by creating a certain types of artificial adversarial examples~\cite{jia2017adversarial,alzantot2018generating,ren2019generating,jin2020bert}, and involving human-and-model-in-the-loop to create dynamic adversarial examples~\cite{nie2019adversarial,wallace2019trick}. 
Further studies discover that a few types of superficial cues (i.e. shortcuts) in the training data, are learned by the models and hence affect the model robustness~\cite{gururangan2018annotation,mccoy2019right,lai2021machine}.
Besides, several studies try to improve the robustness of the neural models by adversarial data augmentation~\cite{min2020syntactic} and data filtering~\cite{le2020adversarial}. 
All these efforts lead us to better find and fix the robustness issues to some extends. 

However, there are several limitations in previous studies. 
First, the analysis and evaluation in previous work focus on just one or a few types of adversarial examples or shortcuts, 
but we need normative evaluation~\cite{linzen2020can, ettinger2020bert,phangadversarially}. 
The goal of the normative evaluation is not to fool a system by exploiting its particular weaknesses, but using systemically controlled datasets to comprehensively evaluate the basic linguistic capabilities of the models in a diverse way. 
Checklist~\cite{ribeiro2020beyond} and Textflint~\cite{gui2021textflint} are great attempts of normative evaluation. However, it is not clear that if the effects of the artificial adversarial methods on artificial examples are still shown on natural texts from real-world applications~\cite{morris2020reevaluating}.
Some other works manually perturb the examples to construct natural examples, but the manual perturbations is time consuming and costly~\cite{gardner2020evaluating}.
Moreover, to the best of our knowledge, there are few Chinese datasets for QM robustness evaluation. 

Towards this end, we create a open-domain Chinese dataset namely \textbf{DuQM} contains natural questions with linguistic perturbation for evaluating the robustness of QM models. 
(1) By \textit{linguistic}, we mean this systematically controlled dataset provides a detailed breakdown of evaluation by linguistic phenomenon. As shown in Tab.~\ref{tab:statistics-exp}, there are 3 categories and 13 subcategories with 32 linguistic perturbation in DuQM, which enables us to evaluate the model performance by each category instead of just a single metric. 
(2) By \textit{natural}, we mean all the questions in DuQM are natural and issued by the users in a commercial search engine. This design can help us to properly evaluate the progress of a model's robustness on natural texts rather than artificial texts which may not preserve semantics and introduce grammatical errors. 

The contributions of this paper can be summarized as follows:
\begin{itemize}[leftmargin=*,noitemsep]
\item We construct a Chinese dataset namely DuQM that contains linguistically perturbed natural questions from a commercial search engine. 
It is a systemically controlled dataset to test the basic linguistic capabilities of the models in a diverse way. (see Sec.~\ref{Constructing QQBugs Dataset} and Sec.~\ref{construction overview})
\item Our experimental results show that 3 characteristics of  DuQM: (1) DuQM is challenging, and has better discrimination power to distinguish the models that perform comparably on other datasets (see Sec.~\ref{sec:challenging}). (2) The detailed breakdown of evaluation by linguistic phenomena in DuQM helps diagnose the advantages and disadvantages of different models (see Sec.~\ref{sec:fine-grained}). 
(3) Extensive experiment shows that the effect of artificial adversarial examples does not work on natural texts of DuQM. DuQM can help us properly evaluate the models' robustness. (see Sec.~\ref{sec:natural}). 
\end{itemize}

The remaining of this paper is organized as follows. Sec.~\ref{Constructing QQBugs Dataset} describes the 3 categories and 13 subcategories with 32 linguistic perturbation in DuQM. Sec.~\ref{Statistics} gives the construction process of DuQM. 
In Sec.~\ref{setup}, we conduct experiments to demonstrate 3 characteristics of DuQM. 
We conclude our work in Sec.~\ref{sec-conclusion}.

\section{Linguistic Perturbations in DuQM}\label{Constructing QQBugs Dataset}

The design of DuQM is aimed at a detailed breakdown of evaluation by linguistic phenomenon. Hence, we create DuQM by introducing a set of linguistic features that we believe are important for model diagnosis in terms of linguistic capabilities. 
Basically, 3 categories of linguistic features are used to build DuQM, i.e., lexical features (see Sec.~\ref{sec-lex}), syntactic features (see Sec.~\ref{sec-syn}), and pragmatic features (see Sec.~\ref{sec-prag}). 
We list 3 categories, 13 subcategories with 32 operations of perturbation in Tab.~\ref{tab:statistics-exp}. The detailed descriptions of all categories are given in this section. 

\begin{table*}
\centering
\newcommand{\tabincell}[2]{\begin{tabular}{@{}#1@{}}#2\end{tabular}}
\resizebox{160mm}{110mm}
{
\small
\begin{tabular}{c|c|c|c|cccccc|l}
\toprule[0.7pt]
  \textbf{ \tabincell{c}{Category}} & \textbf{\tabincell{c}{Subcategory}} & \textbf{\tabincell{c}{Perturbation\\ Operation}} & {\tabincell{c}{\textbf{Label} \\ \#Y / \#N}} &{ \tabincell{c}{BERT\\base}}&{ \tabincell{c}{ERNIE\\base}} & { \tabincell{c}{RoBERTa\\base}}&{ \tabincell{c}{MacBERT\\base}}&{ \tabincell{c}{RoBERTa\\large}}&{ \tabincell{c}{MacBERT\\large}}& \tabincell{c}{Examples and Translation} \\
\midrule[0.7pt]


\multirow{70}*{{\rotatebox{90}{\textbf{Lexical Feature}}}} & \multirow{28}*{\tabincell{c}{Part\\of\\Speech}} & insert n.&
 -/539
&
41.4±3.4 & 40.8±2.1 & \underline{43.0±0.7} & 41.4±2.5 & \textbf{45.4±4.1} & 37.3±2.4 &
 \tabincell{l}{\textbf{E1}: \label{example:E1}鸡蛋怎么炒好吃\;/\; 鸡蛋\colorbox{lightgreen}{\color{darkgreen}面}怎么炒好吃 \\
\quad\;\;\, how to fry eggs\;/\; how to fry egg 
 \colorbox{lightgreen}{\color{darkgreen}noodles}}
 \\\cline{3-11}

 & & insert v. & 
 -/131
 &
 \underline{39.4±0.4} & 33.8±2.6 & 37.4±2.0 & 35.9±2.7 & \textbf{39.9±3.1} & 29.5±3.8 &
 \tabincell{l}{\specialrule{0em}{0em}{1mm}\textbf{E2}: \label{example:E2}梦到西红柿\;/\; 梦到\colorbox{lightgreen}{\color{darkgreen}摘}西红柿\\
\quad\;\;\,dream of tomatoes\;/\; dream of \colorbox{lightgreen}{\color{darkgreen}picking} tomatoes
}
\\\cline{3-11}

 & & insert adj. & 
 -/458
 &
 23.5±1.9 & 19.2±3.7 & \textbf{26.9±4.4} & \underline{23.9±4.2} & 18.1±2.4 & 10.4±2.1 &
 \tabincell{l}{\specialrule{0em}{0em}{1mm}\textbf{E3}: \label{example:E3}有哪些类型的app\;/\; 有哪些类型的\colorbox{lightgreen}{\color{darkgreen}移动}app\\
 \quad\;\;\,what are types of apps\;/\; what are types of \colorbox{lightgreen}{\color{darkgreen}mobile} apps
}
\\\cline{3-11}

 & & insert adv. & 
  -/302
 &
 3.7±0.5 & 4.2±0.5 & 3.8±0.6 & \underline{4.4±1.2} & \textbf{5.8±1.5} & 3.1±1.1 &
 \tabincell{l}{\specialrule{0em}{0em}{1mm}\textbf{E4}: \label{example:E4}为什么打嗝\;/\; 为什么\colorbox{lightgreen}{\color{darkgreen}老}打嗝\\
 \quad\;\;\,why burp\;/\; why \colorbox{lightgreen}{\color{darkgreen}always} burp
}
\\\cline{3-11}

 & & replace n. & 
  -/702
 &
 86.6±0.3 & 86.7±0.1 & 88.3±0.3 & \underline{88.8±1.2} & \textbf{89.4±1.6} &  87.8±0.7&
 \tabincell{l}{\specialrule{0em}{0em}{1mm}\textbf{E5}: \label{example:E5}申请美国\colorbox{lightred}{\color{darkred}绿卡}流程\;/\; 申请美国\colorbox{lightgreen}{\color{darkgreen}签证}流程\\
\quad\;\;\,U.S. \colorbox{lightred}{\color{darkred}green card} application process\;/\; U.S.  \colorbox{lightgreen}{\color{darkgreen}visa} application process
}
\\\cline{3-11}

 & & replace v. & 
  -/466
 &
 71.7±1.1 &77.6±0.8 & 76.9±0.4 & 76.5±1.2 & \underline{81.0±1.6} & \textbf{81.5±2.2} &
 \tabincell{l}{\specialrule{0em}{0em}{1mm}\textbf{E6}: \label{example:E6}为什么\colorbox{lightred}{\color{darkred}下蹲}膝盖疼\;/\; 为什么\colorbox{lightgreen}{\color{darkgreen}下跪}膝盖疼\\
 \quad\;\;\,why knee pain when \colorbox{lightred}{\color{darkred}squatting}\;/\; why knee pain when \colorbox{lightgreen}{\color{darkgreen}kneeling}
}
\\\cline{3-11}

 & & replace adj. & 
  -/472
 &
 74.3±2.1 & 80.0±1.0 & 77.6±0.7 & 81.6±0.5 & \textbf{82.7±1.1} & \underline{82.7±1.6} &
 \tabincell{l}{\specialrule{0em}{0em}{1mm}\textbf{E7}: \label{example:E7}耳朵出血\colorbox{lightred}{\color{darkred}严重}吗\;/\; 耳朵出血\colorbox{lightgreen}{\color{darkgreen}正常}吗\\
 \quad\;\;\,is the ear bleeding \colorbox{lightred}{\color{darkred}serious}\;/\; is the ear bleeding \colorbox{lightgreen}{\color{darkgreen}normal}
}
\\\cline{3-11}

& & replace adv. & 
 -/188
&
19.1±6.1 & 19.3±4.4 & 16.3±3.8 & 23.9±4.6 & \textbf{59.0±4.0} & \underline{56.2±2.0} &
 \tabincell{l}{\specialrule{0em}{0em}{1mm}\textbf{E8}: \label{example:E8}为什么会\colorbox{lightred}{\color{darkred}经常}头晕\;/\; 为什么会\colorbox{lightgreen}{\color{darkgreen}有点}头晕\\
 \quad\;\;\,why \colorbox{lightred}{\color{darkred}regularly} feel dizzy\;/\; why  \colorbox{lightgreen}{\color{darkgreen}slightly} feel dizzy
}
\\\cline{3-11}

 & & replace num.&  
  -/1116
 &
 83.2±1.4 & \underline{91.4±0.4} & 85.9±1.8 & 87.2±0.9 & 88.1±0.5 & \textbf{91.9±1.1} &
 \tabincell{l}{\specialrule{0em}{0em}{1mm}\textbf{E9}: \label{example:E9}血压\colorbox{lightred}{\color{darkred}130}/100高吗\;/\;
 血压\colorbox{lightgreen}{\color{darkgreen}120}/100高吗\\
 \quad\;\;\,is blood pressure \colorbox{lightred}{\color{darkred}130}/100 high \;/\; is blood pressure \colorbox{lightgreen}{\color{darkgreen}120}/100 high}
\\\cline{3-11}

 & & replace quantifier&  
  -/22
 &
 {30.3±6.9} & 25.7±5.2 & 33.3±2.6 & \textbf{34.9±2.6} & 27.3±0.0 & \underline{34.8±10.5} &
 \tabincell{l}{\specialrule{0em}{0em}{1mm}\textbf{E10}: \label{example:E10}一\colorbox{lightred}{\color{darkred}束}花多少钱\;/\;
 一\colorbox{lightgreen}{\color{darkgreen}枝}花多少钱\\
 \qquad\, how much is \colorbox{lightred}{\color{darkred}a bunch of} flower \;/\; how much is  \colorbox{lightgreen}{\color{darkgreen}a}flower}
\\\cline{3-11}

 & & replace phrases &   
  -/197
 &
 \underline{98.0±0.0} & \textbf{98.1±0.2} & 96.6±0.3 & 97.8±0.5 & 97.8±0.2 & 97.5±0 &
 \tabincell{l}{\specialrule{0em}{0em}{1mm}\textbf{E11}: \label{example:E11}如何\colorbox{lightred}{\color{darkred}提高自己的记忆力}\;/\; 如何\colorbox{lightgreen}{\color{darkgreen}增加自己的实力}\\
 \qquad\, how to \colorbox{lightred}{\color{darkred}improve my memory}\;/\; how to \colorbox{lightgreen}{\color{darkgreen}increase my strength}
 }

\\\cline{2-11}

 & \multirow{10}*{\tabincell{c}{Named\\Entity}} & replace loc. & 
  -/458
 &
 \textbf{96.0±0.6} & \underline{95.7±0.2} & 95.4±0.4 & 95.0±0.4 & 94.7±0.4&94.5±0.5 &
 \tabincell{l}{\specialrule{0em}{0em}{1mm}\textbf{E12}: \label{example:E12}\colorbox{lightred}{\color{darkred}山西}春节习俗 \;/\;\colorbox{lightgreen}{\color{darkgreen}陕西}春节习俗 \\
 \qquad\, \colorbox{lightred}{\color{darkred}Shanxi} spring festival customs\;/\; \colorbox{lightgreen}{\color{darkgreen}Shannxi} spring festival customs}
\\\cline{3-11}

 &  & replace org. &  
  -/264
 &
 \textbf{94.9±0.2} & \underline{94.3±0.6} & 91.2±1.4 & 93.4±0.7 & 93.5±0.3 & 93.8±0.1 &
  \tabincell{l}{\specialrule{0em}{0em}{1mm}
  \textbf{E13}: \label{example:E13}\colorbox{lightred}{\color{darkred}北京邮电大学}附近酒店 \;/\; \colorbox{lightgreen}{\color{darkgreen}南京邮电大学}附近酒店\\
 \qquad\, hotels near \colorbox{lightred}{\color{darkred}BUPT}\;/\; hotels near \colorbox{lightgreen}{\color{darkgreen}NJUPT}}
\\\cline{3-11}
 
 &  & replace person &  
  -/468
 &
 90.3±1.3 & 91.0±0.9  & 88.7±1.6 & 91.4±1.6 & \underline{92.3±1.3}& \textbf{93.2±1.1} &
  \tabincell{l}{\specialrule{0em}{0em}{1mm}
  \textbf{E14}: \label{example:E14}\colorbox{lightred}{\color{darkred}陈龙}的妻子 \;/\; \colorbox{lightgreen}{\color{darkgreen}成龙}的妻子\\
 \qquad\, wife of \colorbox{lightred}{\color{darkred}Long Chen}\;/\; wife of \colorbox{lightgreen}{\color{darkgreen}Jackie Chan}}
\\\cline{3-11}
 
 &  & replace product &  
  -/170
 &
 83.7±2.6 &\underline{88.2±2.1} & 82.4±6.9 & 83.3±0.3 & 86.0±1.7 & \textbf{88.8±4.4} &
  \tabincell{l}{\specialrule{0em}{0em}{1mm}
  \textbf{E15}: \label{example:E15}\colorbox{lightred}{\color{darkred}iphone 6}多少钱 \;/\; \colorbox{lightgreen}{\color{darkgreen}iphone6x}多少钱\\
 \qquad\, how much is \colorbox{lightred}{\color{darkred}iphone 6}\;/\; how much is \colorbox{lightgreen}{\color{darkgreen}iphone6x}}
\\\cline{2-11}
 
 & \multirow{10}*{Synonym} & replace n. &  
  405/-
 &
 51.1±1.1 & 59.7±1.3 & 59.7±2.2 & 60.7±2.0 & \underline{63.3±3.1} &\textbf{71.6±4.0} & 
 \tabincell{l}{\specialrule{0em}{0em}{1mm}
 \textbf{E16}: \label{example:E16}\colorbox{lightred}{\color{darkred}猕猴桃}的功效 \;/\;
  \colorbox{lightgreen}{\color{darkgreen}奇异果}的功效\\
\qquad\, health benefits of \colorbox{lightred}{\color{darkred}Chinese gooseberry}\;/\;
health benefits of \colorbox{lightgreen}{\color{darkgreen}Kiwi}}
\\\cline{3-11}

 &  & replace v. &  
  372/-
 &
 80.0±0.9& 81.1±1.6 & 82.5±0.0 & 83.2±1.2 & \underline{84.0±2.0} & \textbf{88.1±1.4} &
  \tabincell{l}{\specialrule{0em}{0em}{1mm}
  \textbf{E17}: \label{example:E17}什么果汁可以\colorbox{lightred}{\color{darkred}减肥} \;/\; 什么果汁可以\colorbox{lightgreen}{\color{darkgreen}减重}\\
 \qquad\, what juice can \colorbox{lightred}{\color{darkred}lose weight}\;/\; what juice can \colorbox{lightgreen}{\color{darkgreen}slim}}
\\\cline{3-11}
 
 &  & replace adj. &  
  453/-
 &
 75.7±1.3 & 77.3±1.1 & 78.8±2.5 & 74.8±0.5 & \underline{79.4±3.4} & \textbf{88.5±1.3} &
  \tabincell{l}{\specialrule{0em}{0em}{1mm}
  \textbf{E18}: \label{example:E18}\colorbox{lightred}{\color{darkred}有趣}搞笑的广告词 \;/\; \colorbox{lightgreen}{\color{darkgreen}幽默}搞笑的广告词\\
 \qquad\, \colorbox{lightred}{\color{darkred}funny} advertising words\;/\; \colorbox{lightgreen}{\color{darkgreen}humerous} advertising words}
\\\cline{3-11}
 
 &  & replace adv. &  
  26/-
 &
 \underline{98.7±2.1} & \textbf{100.0±0.0} & \textbf{100.0±0.0} & \textbf{100.0±0.0} &\textbf{100±0.0} &\textbf{100.0±0.0} & 
  \tabincell{l}{\specialrule{0em}{0em}{1mm}
  \textbf{E19}: \label{example:E19}\colorbox{lightred}{\color{darkred}总是}想睡觉是为什么 \;/\; \colorbox{lightgreen}{\color{darkgreen}老是}想睡觉是为什么\\
 \qquad\, why \colorbox{lightred}{\color{darkred}always} want to sleep \;/\; why \colorbox{lightgreen}{\color{darkgreen}repeatedly} want to sleep}
\\\cline{2-11}

 & Antonym & replace adj. & 
  -/305
 &
 50.6±3.4 &69.6±2.9 & 65.0±1.5 & 73.1±4.3 &\textbf{91.7±2.3} &\underline{90.7±2.3} & 
  \tabincell{l}{\specialrule{0em}{0em}{1mm}
  \textbf{E20}: \label{example:E20}什么水果脂肪\colorbox{lightred}{\color{darkred}低} \;/\; 
  什么水果脂肪\colorbox{lightgreen}{\color{darkgreen}高}\\
  \qquad\, what fruit is  \colorbox{lightred}{\color{darkred}low} in fat\;/\;
  what fruit is \colorbox{lightgreen}{\color{darkgreen}high} in fat
  }
  
\\\cline{2-11}

 & \multirow{6}*{Negation} & negate v. &  
  -/153
 &
 69.9±9.6 & 88.9±1.3 & 84.8±2.9 & \textbf{93.3±1.3} & 88.4±0.9 & \underline{91.4±3.4} & 
 \tabincell{l}{\specialrule{0em}{0em}{1mm}
 \textbf{E21}: \label{example:E21}为什么宝宝哭\;/\; 为什么宝宝\colorbox{lightgreen}{\color{darkgreen}不}哭 \\
 \qquad\, why baby cries\;/\; why baby \colorbox{lightgreen}{\color{darkgreen}doesn't} cry
 }
\\\cline{3-11}
 &  & negate adj. &  
  -/139
 &
 73.1±8.5 & 84.2±1.2 & 82.7±1.4 & \underline{88.0±1.5} & 88.0±2.9 & \textbf{89.4±1.0} & 
  \tabincell{l}{\specialrule{0em}{0em}{1mm}
\textbf{E22}: \label{example:E22}为什么苹果是红的\;/\; 为什么苹果\colorbox{lightgreen}{\color{darkgreen}不是}红的 \\
 \qquad\, why apple is red\;/\; why apple is \colorbox{lightgreen}{\color{darkgreen}not} red }
\\\cline{3-11}
 
 &  & neg.+antonym & 
  59/-
 &
 29.9±2.5 & 34.4±2.5 & 39.0±1.7 & 31.1±2.5 & \underline{40.7±1.7} & \textbf{53.6±0.9} & 
 \tabincell{l}{\specialrule{0em}{0em}{1mm}
 \textbf{E23}: \label{example:E23}\colorbox{lightred}{\color{darkred}激动}怎么办\;/\; \colorbox{lightgreen}{\color{darkgreen}无法}\colorbox{lightgreen}{\color{darkgreen}平静}怎么办 \\
\qquad\, what to do if too \colorbox{lightred}{\color{darkred}excited}\;/\; what to do if \colorbox{lightgreen}{\color{darkgreen}can't}\colorbox{lightgreen}{\color{darkgreen}calm down}
 }
 
\\\cline{2-11}

 & \multirow{4}*{{\tabincell{c}{Temporal\\word}}} & insert & 
  -/120
 &
 26.6±2.1 & 29.1±2.1 & 33.1±0.9 & \underline{41.7±3.3} & \textbf{47.5±5.4} & 33.6±8.5  &
 \tabincell{l}{\specialrule{0em}{0em}{1mm}
 \textbf{E24}: \label{example:E24}北京会下雨吗\;/\; 北京\colorbox{lightgreen}{\color{darkgreen}明天}会下雨吗 \\
 \qquad\, will it rain in Beijing\;/\; will it rain in Beijing\colorbox{lightgreen}{\color{darkgreen}tomorrow}
 }
 \\\cline{3-11}
 & & replace & 
  -/114
 &
 44.1±6.1 & 67.8±2.6 & 55.0±0.5 & 53.8±1.3 & \underline{70.4±6.1} & \textbf{78.6±5.8} &
 \tabincell{l}{\specialrule{0em}{0em}{1mm}
\textbf{E25}: \label{example:E25}\colorbox{lightred}{\color{darkred}昨天}下雪\colorbox{lightred}{\color{darkred}了}吗\;/\; \colorbox{lightgreen}{\color{darkgreen}明儿}会下雪吗 \\
\qquad\, was it \colorbox{lightred}{\color{darkred}snow} yesterday\;/\; will it \colorbox{lightgreen}{\color{darkgreen}snow} tomorrow
 }

 \\\midrule[0.7pt]
 
\multirow{13}*{{\rotatebox{90}{\textbf{Syntactic Feature}}}}
 & Symmetry & swap & 
  533/-
 &
 \underline{97.3±0.4} & \textbf{98.0±0.1} & 95.2±1.7 & 95.9±0.7 & 93.3±0.9 & 92.5±1.9  &
 \tabincell{l}{\textbf{E26}: \label{example:E26}\colorbox{lightred}{
\color{darkred}鱼}和\colorbox{lightred}{\color{darkred}鸡蛋}能一起吃吗\;/\; 
  \colorbox{lightgreen}{\color{darkgreen}鸡蛋}和\colorbox{lightgreen}{\color{darkgreen}鱼}能一起吃吗\\
  \qquad\, can I eat \colorbox{lightred}{\color{darkred}fish} with \colorbox{lightred}{\color{darkred}egg}\;/\; can I eat \colorbox{lightgreen}{\color{darkgreen}egg} with \colorbox{lightgreen}{\color{darkgreen}fish}
}
\\\cline{2-11}
 
  & Asymmetry & swap & 
   -/497
  &
  14.5±2.0 & 18.3±3.7 & 26.8±3.2 & 26.4±2.5 & \textbf{52.0±4.6} & \underline{49.1±10.8} &
  \tabincell{l}{\specialrule{0em}{0em}{1mm}
  \textbf{E27}: \label{example:E27}\colorbox{lightred}{\color{darkred}北京}到\colorbox{lightred}{\color{darkred}上海}航班\;/\; 
  \colorbox{lightgreen}{\color{darkgreen}上海}到\colorbox{lightgreen}{\color{darkgreen}北京}航班\\
  \qquad\, \colorbox{lightred}{\color{darkred}Beijing} to \colorbox{lightred}{\color{darkred}Shanghai} flights\;/\;
  \colorbox{lightgreen}{\color{darkgreen}Shanghai} to \colorbox{lightgreen}{\color{darkgreen}Beijing} flights
}
\\\cline{2-11}
&\multirow{5}*{\tabincell{c}{Negative\\Asymmetry}}&&&&\\
 &  & swap + negate &
 49/-
 &
  \textbf{47.6±3.4}& 37.4±7.7 & \underline{44.2±1.1} & 25.8±3.1 & 23.1±6.7 & 29.9±1.9 &
 \tabincell{l}{
  \textbf{E28}: \label{example:E28}\colorbox{lightred}{\color{darkred}男人}比\colorbox{lightred}{\color{darkred}女人}更\colorbox{lightred}{\color{darkred}高}吗\;/\; 
  \colorbox{lightgreen}{\color{darkgreen}女人}比\colorbox{lightgreen}{\color{darkgreen}男人}更\colorbox{lightgreen}{\color{darkgreen}矮}吗\\
  \qquad\, are \colorbox{lightred}{\color{darkred}men}\colorbox{lightred}{\color{darkred}taller}  than \colorbox{lightred}{\color{darkred}women}\;/\; are \colorbox{lightgreen}{\color{darkgreen}women}\colorbox{lightgreen}{\color{darkgreen}shorter} than \colorbox{lightgreen}{\color{darkgreen}men}
}
\\
 &&&&&&&&\\\cline{2-11}
 
 & Voice & insert passive word &
  94/37
 &
 76.8±1.4& 72.5±0.0 & \underline{77.4±0.9} & 74.0±0.7 & \textbf{85.2±1.4} & 74.8±2.2 &
 \tabincell{l}{
 \specialrule{0em}{0em}{1mm}
  \textbf{E29}: \label{example:E29}梦见狗咬左腿\;/\;梦见\colorbox{lightgreen}{\color{darkgreen}被}狗咬左腿\\
 \qquad\, dreamed of being bitten by a dog\;/\; 
 dreamed of being bitten by a dog
 } 
\\
\midrule[0.7pt]
 
\multirow{7}*{\rotatebox{90}{\textbf{{Pragmatic Feature}}}}
 & Misspelling & replace &
 468/-
 &
 \textbf{68.0±2.0} &\underline{65.1±0.2} & 64.2±0.6 & 65.0±2.3 & 63.5±1.8 & 63.2±1.6 &
 \tabincell{l}{
  \textbf{E30}: \label{example:E30}什么\colorbox{lightred}{\color{darkred}纹身}适合我\;/\;什么\colorbox{lightgreen}{\color{darkgreen}文身}适合我
\\ 
\qquad\, what \colorbox{lightred}{\color{darkred}tattoo} suits me\;/\; what \colorbox{lightgreen}{\color{darkgreen}tatoo} suits me
}\\
\cline{2-11}
& \multirow{2}*{\tabincell{c}{Discourse Particle\\(Simple)}} & insert or replace &
213/-
&
98.7±0.5 & 98.4±0.2 & 98.6±0.5 & 99.2±0.2 & \underline{99.5±0.0} & \textbf{99.8±0.2} &
\tabincell{l}{
 \textbf{E31}: \label{example:E31}人为什么做梦 \;/\;  
\colorbox{lightgreen}{\color{darkgreen}那么}人为什么做梦 \\ 
\qquad\, why people dream \;/\;  
\colorbox{lightgreen}{\color{darkgreen}so} why people dream}\\
\specialrule{0em}{0em}{1mm}
\cline{2-11}
&  \multirow{2}*{\tabincell{c}{Discourse Particle\\(Complex)}} & insert or replace &
131/-
&
46.5±0.6 & 56.2±2.0 & 64.1±2.0 & 61.6±1.6 & \underline{65.1±3.4} & \textbf{68.4±0.3} &
\tabincell{l}{\specialrule{0em}{0em}{1mm}
 \textbf{E32}: \label{example:E32}附近最好的餐厅 \;/\;  
\colorbox{lightgreen}{\color{darkgreen}求助我旁边}哪家餐厅\colorbox{lightgreen}{\color{darkgreen}最好吃}? \\ 
\qquad\, best restaurant nearby \;/\;  
\colorbox{lightgreen}{\color{darkgreen}heeelp!!!}which restaurant is best \colorbox{lightgreen}{\color{darkgreen}in my area}?}



 \\
\midrule[0.7pt]
\textbf{Total} & \textbf{13} & \textbf{32} & \textbf{2803/7318} & \multicolumn{6}{l|}{-}  & -  \\
\bottomrule[0.7pt]

\end{tabular}
}
\caption{Categories of DuQM (described in Sec. \ref{Constructing QQBugs Dataset}) and performance of 6 models on DuQM (discussed in Sec. \ref{setup}).
 \textbf{Bold face} and \underline{underlined} indicate the first and second highest accuracy for each testing scenario.}\label{tab:statistics-exp}
\end{table*}

\subsection{Lexical Features}\label{sec-lex}
Lexical features are associated with vocabulary items, i.e. words. As a word is the smallest independent but meaningful unit of speech
, an operation on a single word may change the meaning of the entire sentence. It is a basic but crucial capability for models to understand word and perceive word-level perturbations. To provide a fine-grained evaluation for model's capability of lexical understanding, we further consider 6
subcategories:

\noindent\textbf{Part of Speech.}
Parts of speech (POS), or word classes, describe the part a word plays in a sentence. DuQM considers 6 POS in Chinese grammar, including noun, verb, adjective, adverb, numeral and quantifier, which are content words that carry most meaning of a sentence. In this subcategory, we aim to test 
the models' understanding of related but not identical words with different POSs.
As the example 1 in Tab.~\ref{tab:statistics-exp}~\footnote{All examples discussed in this section are presented in Column \textit{Example and Translation} of Tab.~\ref{tab:statistics-exp}.} shows, inserting only one noun "\textit{面~noodles}" makes the sentence meaning different.
Furthermore, in this subcategory we provides a set of examples focusing on phrase-level perturbations to check model's capability on understanding word groups that act collectively as a single part of speech (see example 11).

\noindent\textbf{Named Entity.}
Different from common nouns that refer to generic things, a named entity (NE) is a proper noun which refers to a specific real-world object. The close relation to world knowledge makes NE ideal for observing models' understanding of the meaning of names and background knowledge about entities. In DuQM, we include \textit{Named Entity} an independent subcategory to test the model's behavior of named entity recognition, and focus on 4 types of NE most commonly seen, i.e., location, organization, person and product. 
Example 12 is a search query and its perturbation on NE.
The two named entities, \textit{"山西~Shanxi"} and \textit{"陕西~Shaanxi"}, are similar at character level but denote two different locations. 
We expect that the models can capture the subtle difference.

\noindent\textbf{Synonym.}
A synonym is a word or phrase that means exactly or nearly the same as another word or phrase in a given language. This subcategory aims to test whether models can identify two semantically equivalent questions whose surface forms only differ in a pair of synonyms. As in example 16, 
the two sentences differ only in two words, both of which refer to Kiwifruit, so they have the same meaning. 

\noindent\textbf{Antonym.}
In contrast to synonyms, antonyms are words within an inherently incompatible binary relationship. This subcategory examines model's capability on distinguishing words with opposed meanings. We mainly focus on adjective's opposite, e.g., "高\textit{high}" and "低\textit{low}" (see example 20).

\noindent\textbf{Negation.}
Negation is another way to express contradiction. To negate a verb or an adjective in Chinese, we normally put a negative before it, e.g., "不\textit{not}" before "哭\textit{cry}" (example 21), "不是\textit{not}" before "红的\textit{red}" (example 22). The negative before the verb or the adjective negates the statement.
It is an effective way to analyze model's basic skill of figuring out the contradictory meanings even there is only a minor change. 

Moreover, we include some equivalent paraphrases with negation in this subcategory. In example 23, "无法平静\textit{can't calm down}" is the negative paraphrase of "激动\textit{excited}", so that the paraphrase sentence is equivalent to the positive sentence.
We believe that a robust QM system should be able to recognize this kind of paraphrase question pairs.

\noindent\textbf{Temporal Word.}
Temporal reasoning is the relatively higher-level linguistic capability that allows the model to reason about a mathematical timeline. Unlike English, verbs in Chinese do not have morphological inflections. Tenses and aspects are expressed either by temporal noun phrases like "明天\textit{tomorrow}" (examples 24) or by aspect particles like "了\textit{le}", which indicates the completion of an action (examples 25). This subcategory focuses on the temporal distinctions and helps us evaluate the models' temporal reasoning capability. 

\subsection{Syntactic Features}\label{sec-syn}
While single word sense is important to question meaning, how words composed together into a whole also affects sentence understanding. We believe the the relations among words in a sentence is important information for models to capture, so we focus on several types of syntactic features in this category. We pre-define 4 linguistic phenomena that we believe is meaningful to locate model's strength and weakness, and describe them here.

\noindent\textbf{Symmetry.}
Sometimes paraphrases can be generated by only swapping the two conjuncts around in a structure of coordination. As shown in example 26, "鱼\textit{fish}" and "鸡蛋\textit{egg}" are joined together by the conjunction "和\textit{and}", which have the symmetric relation to each other. Even if we swap them around, the sentence meaning will not change. We name this subcategory \textit{Symmetry}, with which we aim to explore if a model captures the inherent dependency relationship between words.

\noindent\textbf{Asymmetry.}
Some words (such as "和and") denote symmetric relations, while others (for example, preposition "到to") denote asymmetric. Example 27 shows a sentence pair in which the word before the preposition "到\textit{to}" is an adverbial and the word after it is the object. Swapping around the adverbial and the object of the prepositional phrase will definitely leads to a nonequivalent meaning. If a model performs well only on subcategory \textit{Symmetry} or \textit{Asymmetry}, it may rely on shortcuts instead of the understanding of the syntactic information.

\noindent\textbf{Negative Asymmetry.}
To further explore the syntactic capability of QM model, DuQM includes a set of test examples which consider both syntactic asymmetry and antonym, and we name this category \textit{Negative Asymmetry}. In example 28, the asymmetric relation between "男人\textit{men}" and "女人\textit{women}" and the opposite meaning of "高\textit{taller}" and "矮\textit{shorter}" resolve to an equivalent meaning.
With this subcategory, we can better explore model's capability of inferring more complex syntactic structure.

\begin{figure}
\centering
\subfigure[Active voice question.]{
    \label{subfig:parser-a}
  \includegraphics[scale=0.8]{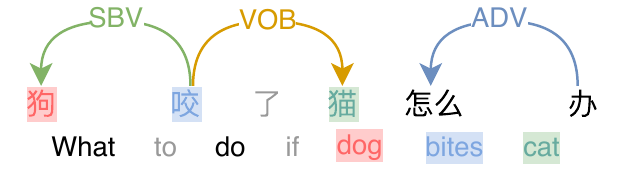}}

\subfigure[Passive voice paraphrase question.]{
    \label{subfig:parser-b}
  \includegraphics[scale=0.8]{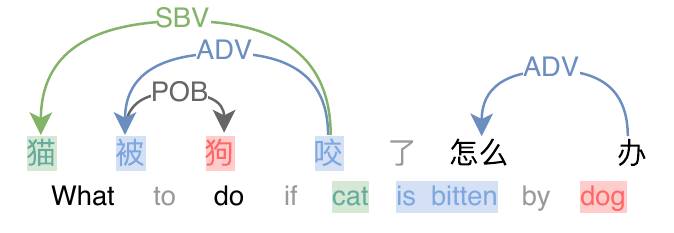}}
  
\subfigure[Passive voice non-paraphrase question.]{
    \label{subfig:parser-c}
  \includegraphics[scale=0.8]{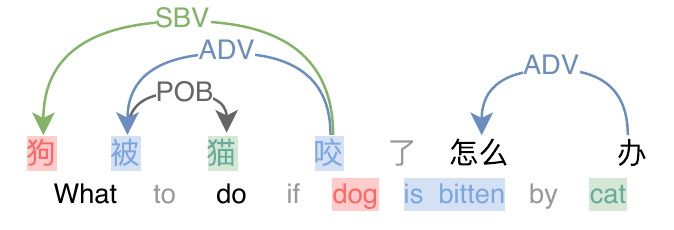}}
  
\caption{The dependency relations of active voice and passive voice questions. 
}
\label{fig:parser-fig}
\end{figure}

\noindent\textbf{Voice.}
Another crucial syntactic capability of models is to differentiate active and passive voices. In Chinese, the most common way to express the passive voice is using Bei-constructions which feature an agentive case marker "被\textit{bei}". The subject of a Bei-construction is the patient of an action, and the object of the preposition "被\textit{bei}" is the agent. Compared to Fig.\ref{subfig:parser-a}, the additional "被\textit{bei}" and the change of word order of "猫\textit{cat}" and "狗\textit{dog}" in Fig.\ref{subfig:parser-b} convert the sentence from active to passive voice, but the two sentences have the same meaning. If we further change the word order from Fig.\ref{subfig:parser-b} to Fig.\ref{subfig:parser-c}, the sentence still uses passive voice but has different meaning.

Passive voice is not always expressed with an overt "被\textit{bei}". Sometimes a sentence without any passive marker is still in passive voice. In example 29, although the first sentence is without "被\textit{bei}", it expresses the same meaning as the second one. 
There are a set of active-passive examples in this category, which are effective to evaluate model's performance on active and passive voices.

\subsection{Pragmatic Features}\label{sec-prag}
Lexical items ordered by syntactic rules are not all that make a sentence mean what it means.  Context, or the communicative situation that influence language use, has a part to play. We include some pragmatic features in DuQM so as to observe whether models are able to understand the contextual meaning of sentences.

\noindent\textbf{Misspelling.}
Misspellings are quite often seen by search engines and question-answering systems, which are mostly unintentional. Models should have the capability to capture the true intention of the questions with spelling errors to ensure the robustness. In example 30, despite the misspelled word "文身\textit{tatoo}" the two questions mean the same,
In some real world situations, models should understand misspellings appropriately. For example, when users search a query but type in misspelling, a robust model will still give the correct result.

\noindent\textbf{Discourse Particle.}
Discourse particles are words and small expressions that contribute little to the information the sentence convey, but play some pragmatic functions such as showing politeness, drawing attention, smoothing utterance, etc. As in example 32, the word "求助\textit{help}" is used to draw attention and bring no additional information to the sentence.
Whether using these little words do not change the sentence meaning. It is necessary to a model to identify the semantic equivalency when such words are used.

\section{Construction}\label{Statistics}\label{construction overview}

\begin{figure}
\centering
  \includegraphics[scale=0.80]{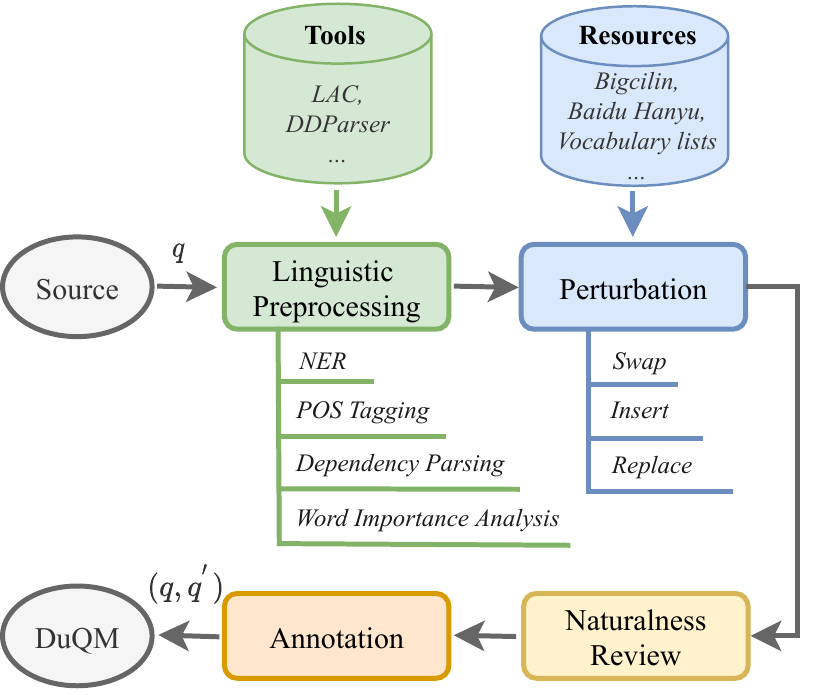}
\caption{Construction process of DuQM.}
\label{fig:construction}
\end{figure}
We design DuQM as a \textit{diverse} and \textit{natural} corpus. The construction process of DuQM is divided into 4 steps and illustrated in Fig.~\ref{fig:construction}. Firstly, we preprocess the source questions to obtain their linguistic knowledge, which will be used to perturb the source texts. Then we pair the source and perturbed question as an example. The examples' naturalness is reviewed by human evaluators. At last, the examples are annotated manually and DuQM is finally constructed. 
We introduce the construction details in the following:

\noindent
\textbf{Linguistic Preprocessing.}
We collect a large number of source questions from the search query log of a commercial search engine. All the source questions are natural and then we perform several linguistic preprocessings on them: named entity recognition, POS tagging, dependency parsing, and word importance analysis.
The linguistic knowledge about the source questions we obtained in this step will be used for perturbation.

\noindent
\textbf{Perturbation.}
We conduct different perturbation operations for different subcategories. In general, we perturb the sentences in 3 ways:
\begin{itemize}[leftmargin=*,noitemsep]
\item \textbf{replace}: replace a word with another word, e.g., for category \textit{Synonym}, we replace one word with its synonym;
\item \textbf{insert}
: insert an additional word, e.g., for category \textit{Temporal word}, we insert temporal word to the source question;
\item \textbf{swap}: swap two words. This operation is only used in \textit{Syntactic Feature}.
\end{itemize}

The perturbations of all categories are listed in column \textit{Perturbation Operation} of Tab.~\ref{tab:statistics-exp}, and the perturbation details will be given in Appendix~\ref{construction details}.

\noindent
\textbf{Naturalness Review.}
To ensure the generated sentences are natural, we examine their appearances in the search log and only retain the sentences which have been entered into the search engine.

\noindent\textbf{Annotation.}
The source question and generated question are paired together as an example. Then the examples are evaluated by evaluators from our internal data team. They need to evaluate whether the examples are fluent, grammatically correct, and correctly categorized. The low-quality examples are discarded and the examples with inappropriate categories are re-classified. 

Then the question pairs are annotated by the linguistic experts from our internal data team. Semantically equivalent question pairs are positive examples, and inequivalent pairs are negative. 
The annotators are required a approval rate higher than 99\% for at least 1,000 prior tasks. 
Each example is annotated by three annotators, and the examples will be tagged with the label which more than 2 annotators choose. To further ensure the annotation quality, 10\% of the annotated examples are selected randomly and reviewed by another senior linguistic expert and and if the review accuracy is lower than 95\%, the annotation linguistic experts need to re-annotate all the examples until the accuracy is higher than 95\%. Since all annotators are linguistic experts from our internal data team instead of crowd-sourcing, we do not need to use IAA to measure the annotation quality. Generally, only 0.002\% (20/10,167) generated examples are not fluent or not grammatically correct, and only 0.02\% (195/10,121) generated examples are re-annotated manually. The overall annotation process is illustrated in Fig.~\ref{fig:expert_eval}. 

\begin{table}
\begin{spacing}{1.4}
\scriptsize
\centering
\newcommand{\tabincell}[2]{\begin{tabular}{@{}#1@{}}#2\end{tabular}}
\setlength{\tabcolsep}{2mm}{
\begin{tabular}{c|cc|ccc}
\toprule[0.7pt]
\multirow{2}{*}{\tabincell{c}{\textbf{Category}}} & \multicolumn{2}{c|}{\textbf{Length}} & \multicolumn{3}{c}{\textbf{\#}}  \\
  & q & q' & Y  & N  & All  \\
\midrule[0.7pt]
{Lexical}  & 8.58 & 8.89  & 1,315 & 6,784  & 8,099               \\
{Syntactic}  & 9.86 & 9.89  & 678    & 532  & 1,210               \\
{Pragmatic}   & 8.73    & 9.03    & 812  & 0 & 812                \\
\midrule[0.7pt]
\textbf{Avg / Total}   & 8.74    & 8.90    & 2,805   & 7,316   & 1,0121              \\
\bottomrule[0.7pt]
\end{tabular}}
\caption{Data statistics of DuQM.}\label{tab:statistics-CQM_robust}
\end{spacing}
\end{table}

Eventually, we generate 10,121 examples for DuQM. The class distribution of all categories are given in Tab.~\ref{tab:statistics-exp}. Additional data statistics are provided in Tab.~\ref{tab:statistics-CQM_robust}.

\section{Experiments}\label{setup}
In this section, we conduct experiments to discuss 3 characteristics (char.) of DuQM. In Sec.~\ref{sec:experimental setup}, we provide the experimental setup and the evaluation metrics. 
In Sec.~\ref{sec:challenging}, Sec.~\ref{sec:fine-grained} and Sec.~\ref{sec:natural}, we give the experimental results and discussions.

\subsection{Experimental Setup}\label{sec:experimental setup}
\paragraph{Datasets.}
To evaluate the robustness of QM models, we select LCQMC to fine-tune the models and evaluate the models' performance on DuQM. LCQMC is a large-scale Chinese QM corpus proposed by Harbin Institute of Technology in \textit{\textbf{general domain}} and the source questions are collected from Baidu Knows (a popular Chinese community question answering website), which \textit{\textbf{are similar to the search queries in form}}. Specifically, we firstly fine-tune the models on \textit{LCQMC\textsubscript{train}}. Then we choose the model with the best performance on \textit{LCQMC\textsubscript{dev}} and report the results of the chosen models on \textit{LCQMC\textsubscript{test}} and DuQM.

And it is worth mentioning that LCQMC is in general domain and its source questions are similar to the search query, which are the form of source questions for DuQM. In other words, \textbf{DuQM is not a out-of-domain (ood) test set of LCQMC}, so that the models' low performance on DuQM could not be attributed to being ood.
Tab.~\ref{tab:statistics-LCQMC} presents the statistics of LCQMC.
\begin{table}
\begin{spacing}{1.4}
\centering
\scriptsize
\newcommand{\tabincell}[2]{\begin{tabular}{@{}#1@{}}#2\end{tabular}}
\begin{tabular}{c|c|c|c}
\toprule[0.7pt]
\textbf{Model} & LCQMC\textsubscript{test} & DuQM &$\triangle$ \\
\midrule[0.7pt]
BERT\textsubscript{b} &87.1±0.1 &66.6±0.6 & -20.5\\
ERNIE\textsubscript{b} & 87.3±0.1& 69.8±0.3 &-17.5 \\
RoBERTa\textsubscript{b} & 87.2±0.4&69.5±0.1 &-17.7\\
MacBERT\textsubscript{b} &87.4±0.3 &70.3±0.6 &-17.1\\
\hline
RoBERTa\textsubscript{l} & \textbf{87.7±0.1}&\textbf{73.8±0.3} &-13.9\\
MacBERT\textsubscript{l} &\underline{87.6±0.1} &\underline{73.8±0.5} &-13.8\\
\bottomrule[0.7pt]
\end{tabular}
\caption{Accuracy(\%) on LCQMC\textsubscript{test} and DuQM. \textsubscript{b} indicates base, and \textsubscript{l} indicates large.}\label{tab:main-results}
\end{spacing}
\end{table}

\paragraph{Models.}
We choose 6 popular pre-trained models to conduct experiments:
\textit{BERT}\textsubscript{b}~\citep{devlin2018bert}, \textit{ERNIE}\textsubscript{b}~\citep{sun2019ernie}, \textit{RoBERTa}\textsubscript{b}, \textit{RoBERTa}\textsubscript{l}~\citep{liu2019roberta}, \textit{MacBERT}\textsubscript{b}, \textit{MacBERT}\textsubscript{l}~\citep{cui2020revisiting}. A detailed comparison is provided in Tab.~\ref{tab:pretrain_model_description} (in Appendix). 

\paragraph{Evaluation Metrics.}\label{sec:metrics}
QM problem is normally formulated as a binary classification task. Like most classification tasks, we use accuracy to evaluate a single model's performance, which is the proportion of correct predictions among the total number of the examined examples. As \textit{DuQM} is a fine-grained corpus consisting of a set of linguistic categories and each category differs in size, 
we use the \textit{\textbf{micro-averaged}} and the \textit{\textbf{macro-averaged accuracy}} to compare the models' performances on DuQM, which can help us better indicate the models' ability on different categories.

Training details about our experiments are described in Appendix~\ref{appendix:train details}.
\begin{table*}
\begin{spacing}{1.4}
\scriptsize
\centering
\newcommand{\tabincell}[2]{\begin{tabular}{@{}#1@{}}#2\end{tabular}}
\begin{tabular}{cc|cccccc|c|c|c|c}
\toprule[0.7pt]
\multicolumn{2}{c|}{\multirow{2}*{\textbf{\tabincell{c}{Models}}} }
 & \multicolumn{7}{c|}{\textbf{Lexical}} & \multirow{2}*{\textbf{Syntactic}} & \multirow{2}*{\textbf{Pragmatic}} & \multirow{2}*{\textbf{DuQM}} \\
\multicolumn{2}{c|}{ }& POS & NE & Synonym & Antonym & Negation & Temporal & \textbf{Lexical} &  &  & \\
 \midrule[0.7pt]
\multirow{2}*{\tabincell{c}{BERT\textsubscript{b}}} & micro & 62.1±1.1 & 92.3±0.5 & 69.5±0.4 & 50.6±3.4 &
64.4±5.9 & 35.1±3.3 & 67.2±0.7 & 59.1±0.4 &
72.6±1.6 & 66.6±0.6  \\
& macro & 51.9±1.5 & 91.2±0.7 & 76.4±0.6 & 50.6±3.4 &
57.6±4.4 & 35.5±3.3 & 61.4±1.2 & 59.1±0.7 &
71.1±1.1 & 62.0±0.9  \\\hline

\multirow{2}*{\tabincell{c}{ERNIE\textsubscript{b}}} & micro & 64.6±0.5 & \underline{92.8±0.4} & 73.2±0.9 & 69.6±2.9 &
77.8±1.1 & 48.0±1.9 & 71.0±0.3 & 60.0±1.2 &
72.4±0.3 & 69.8±0.3  \\
& macro & 52.4±0.7 & \underline{92.3±0.6} & 79.5±0.7 & 69.6±2.9 &
69.1±1.2 & 48.5±1.9 & 65.5±0.5 & 56.5±1.0 &
73.2±0.8 & 65.1±0.3  \\\hline

\multirow{2}*{\tabincell{c}{RoBERTa\textsubscript{b}}} & micro & 64.2±0.1 & 90.6±1.8 & 74.2±1.4  &65.0±1.5 & 76.3±1.7 & 43.7±0.2 &
70.1±0.1 & 63.1±0.6 & 73.3±0.1 & 69.5±0.1  \\
& macro & 53.3±0.2 & 89.4±2.5 & 80.3±1.1 & 65.0±1.5 & 68.8±1.3 & 44.0±0.2 &
65.0±0.1 & 60.9±0.6 & 75.6±0.5 & 65.5±0.1 \\\hline

\multirow{2}*{\tabincell{c}{MacBERT\textsubscript{b}}} & micro & 64.8±1.1 & 92.0±0.7 & 73.3±1.1  & 73.1±4.3 & \underline{80.7±0.5} & 47.6±1.3 &
71.2±0.7 & 62.1±1.0 & \underline{73.4±1.5} & 70.3±0.6  \\
& macro & 54.2±0.9 & 90.7±0.6 & 79.7±0.5 & 73.1±4.6 & 70.7±0.1 & 47.7±0.2 &
66.3±0.2 & 55.5±0.7 & 75.2±1.1 & 65.8±0.1  \\\hline

\multirow{2}*{\tabincell{c}{RoBERTa\textsubscript{l}}} & micro & \textbf{67.2±0.9} &
92.5±0.3 & \underline{76.0±2.1} & \textbf{91.7±2.3} & 80.2±0.8 & \textbf{58.6±2.8} &
\underline{74.1±0.3} &
\textbf{72.6±1.4} &
73.2±1.9 &
\textbf{73.8±0.3}  \\
& macro & \textbf{57.7±0.6} &
91.6±0.3 & \underline{81.7±1.6} & \textbf{91.7±2.3} & \underline{72.3±0.6} & \textbf{59.0±2.7} &
\underline{70.2±0.3} &
\textbf{63.4±1.2} &
\underline{76.0±2.0} & \underline{69.8±0.2}  \\\hline

\multirow{2}*{\tabincell{c}{MacBERT\textsubscript{l}}} & micro & \underline{65.6±0.8} &
\textbf{93.2±0.6} & \textbf{83.2±1.6} & \underline{90.7±2.3} & \textbf{84.3±1.3} & \underline{55.5±4.0} &
\textbf{74.4±0.4} &
\underline{70.2±3.7} &
\textbf{73.7±1.1} &
\underline{73.8±0.5}  \\
& macro & \underline{54.7±0.9} &
\textbf{92.6±0.9} & \textbf{87.1±1.2} & \underline{90.7±2.3} & \textbf{78.1±0.9} & \underline{56.1±4.0} &
\textbf{70.7±0.5} &
\underline{61.6±2.4} &
\textbf{77.1±0.6} & \textbf{70.2±0.5}  \\

\bottomrule[0.7pt]

\end{tabular}
\caption{The micro-averaged and macro-averaged accuracy are on each category of DuQM.}\label{tab:exp-micro-macro}
\end{spacing}
\end{table*}

\subsection{Char. 1: Challenging and with Better Discrimination Ability}\label{sec:challenging}
Tab.~\ref{tab:main-results} shows the performances of models on held-out set LCQMC\textsubscript{test} and our DuQM, which presents the primary characteristics of DuQM:

\noindent \textbf{Challenging.} Comparing to the \textit{held-out test} on LCQMC\textsubscript{test}, all models achieve lower performance on DuQM.
As shown in Tab.~\ref{tab:main-results}, all models achieve accuracy higher than $87\%$ on LCQMC\textsubscript{test}, but show a significant performance drop on DuQM. Column $\triangle$ in Tab.~\ref{tab:main-results} shows the differences between models' performances on LCQMC\textsubscript{test} and DuQM, which presents that the performance on DuQM is lower than on LCQMC\textsubscript{test} by at most 20.5\%. DuQM is more \textbf{challenging}, and it can better reflect true capability of QM models. 
\begin{table}
\begin{spacing}{1.4}
\centering
\scriptsize
\newcommand{\tabincell}[2]{\begin{tabular}{@{}#1@{}}#2\end{tabular}}
\setlength{\tabcolsep}{1.2mm}{
\begin{tabular}{l|cc|cc|c}
\toprule
&PWWS&PWWS\textsubscript{nat}&FOOLER&FOOLER\textsubscript{nat}&CHECKLIST\textsubscript{nat}\\
\midrule[0.7pt]
Train&159,503&-& 64,086&-&\\
Test&400&200&400&200&400 \\
\bottomrule[0.7pt]
\end{tabular}
}
\caption{Statistics of the adversarial examples.}\label{tab:adv-statistics-results}
\end{spacing}
\end{table}

\begin{table*}
\begin{spacing}{1.4}
\scriptsize
\centering
\newcommand{\tabincell}[2]{\begin{tabular}{@{}#1@{}}#2\end{tabular}}
\begin{tabular}{lc|cc|cc|c|cc}
\toprule[0.7pt]
 \multirow{2}*{\textbf{Training set}} & \multirow{2}*{\textbf{LCQMC}} & \multicolumn{4}{c|}{\textbf{Attack test set}} & \multirow{2}*{\textbf{CHECKLIST\textsubscript{nat}}} & \multicolumn{2}{c}{\textbf{DuQM}} \\
 & & PWWS & PWWS\textsubscript{nat}& FOOLER & FOOLER\textsubscript{nat}& & Micro & Macro\\
\midrule[0.7pt]

 LCQMC & \underline{87.7} & 58.1 & 81.5& 57.1 & 87.8& \underline{76.9} & \underline{73.8} & 69.8 \\


 LCQMC+PWWS & \underline{87.7}\textcolor{textgreen}{\textsubscript{+0.0}} & \textbf{97.6}\textcolor{textgreen}{\textsubscript{+39.5}} & 81.8\textcolor{textgreen}{\textsubscript{+0.3}}& \underline{73.1}\textcolor{textgreen}{\textsubscript{+16.0}} & 87.6\textcolor{textred}{\textsubscript{-0.2}}& 76.0\textcolor{textred}{\textsubscript{-0.9}} & \textbf{75.2}
\textcolor{textgreen}{\textsubscript{+1.4}} & \underline{70.4}\textcolor{textgreen}{\textsubscript{+0.6}} \\

 LCQMC+FOOLER & 87.5\textcolor{textred}{\textsubscript{-0.2}} & \underline{78.5}\textcolor{textgreen}{\textsubscript{+20.4}} & \textbf{83.8}\textcolor{textgreen}{\textsubscript{+2.3}}& \textbf{80.8}\textcolor{textgreen}{\textsubscript{+23.7}} & 82.0\textcolor{textred}{\textsubscript{-5.8}}&\textbf{79.2}\textcolor{textgreen}{\textsubscript{+2.3}} & 71.4
\textcolor{textred}{\textsubscript{-2.4}} & 68.8\textcolor{textred}{\textsubscript{-1.0}} \\



\bottomrule[0.7pt]
\end{tabular}
\caption{
Adversarial training results of RoBERTa\textsubscript{l}. 'FOOLER' refers to 'TEXTFOOLER'. We use green and red subscripts to represent a higher and lower accuracy respectively.
} \label{tab:adv-results}
\end{spacing}
\end{table*}


\noindent \textbf{Better Discrimination Ability.} 
DuQM can better distinguish the models' performances. As shown in Tab.~\ref{tab:main-results}, all the models have similar performances on LCQMC\textsubscript{test} (around 87\%), but different performances on DuQM: the accuracy of base models differ from 66.6\% to 70.3\%, and the large models show higher performance (73.8\%). In conclude, DuQM shows a better discrimination ability to evaluate models.

It demonstrates that DuQM can better evaluate the robustness of QM models.
\subsection{Char. 2: Diagnose Model in Diverse Way}\label{sec:fine-grained}
DuQM corpus is a fine-grained corpus which has 3 linguistic categories and 13 subcategories and enables a detailed breakdown of evaluation on different linguistic phenomena. In Tab.~\ref{tab:statistics-exp} we give the performances of 6 models on all fine-grained categories of DuQM, and Tab.~\ref{tab:exp-micro-macro} reports the micro-averaged and macro-averaged accuracy. By comparing these results, we introduce the second characteristic of DuQM: it can diagnose the strengths and weaknesses of the models in a diverse way. Several interesting observations are noticed:
(from Tab.~\ref{tab:statistics-exp} and ~\ref{tab:exp-micro-macro}):
\begin{itemize}[leftmargin=*,noitemsep,topsep=0pt]
    \item [1)] \textit{In most categories}, large models 
    outperform base models. As the large models have more parameters and larger pre-training corpus, it is reasonable that they have better capabilities than relatively smaller models.
    \item [2)] In \emph{Named Entity}, all models show good performances (higher than 90\%). Another interesting finding is that although ERNIE\textsubscript{b} is a relatively small model, it performs slightly better than RoBERTa\textsubscript{l} on this subcategory, which might attribute to the entity masking strategy for pre-training.
    \item [3)] MacBERT\textsubscript{l} is significantly better than others in \emph{Synonym}. We suppose that it benefits from using similar words instead of random words for masking when pre-training. Moreover, RoBERTa\textsubscript{l} and MacBERT\textsubscript{l} have remarkable better performance in \emph{Antonym}.
    \item [4)] The overall low performances in \textit{Temporal word} represent that all models lack the capability of temporal reasoning.
    \item [5)] All models have surprisingly poor performances on \emph{Asymmetry} while good performances in \emph{Symmetry}.
    We suppose that lack of learning word orders would result in a wrong prediction when the words orders are altered.
    \item [6)] BERT\textsubscript{b} and ERNIE\textsubscript{b} perform better on \emph{Misspelling}, and RoBERTa\textsubscript{b} and MacBERT\textsubscript{b} are relatively better on \emph{Complex Discourse Particles}.
\end{itemize}

\noindent In general, DuQM diagnoses models from a linguistic perspective and can help us identify the strengths and weaknesses of the models.
\subsection{Char. 3: Natural Adversarial Examples}\label{sec:natural}
DuQM is a dataset generating by linguistically perturbing natural questions. We argue that this kind of natural adversarial examples is beneficial to a \textit{robustness evaluation}.
To prove that, we conduct an experiment to compare the performances of 2 adversarial training (AT) methods PWWS~\cite{ren2019generating} and TextFooler~\cite{jin2020bert} on artificial and natural test examples:
\begin{itemize}[leftmargin=*,noitemsep,topsep=0pt]
\item {\textit{Artificial examples}},  which are generated \emph{artificially} and may not preserve semantics and introduce grammatical errors. We employ 2 methods PWWS and TextFooler on LCQMC\textsubscript{test} to generate artificial adversarial examples. These two methods generate adversarial examples by replacing words with synonyms until models are fooled.
\item {\textit{Natural examples}} are texts within linguistic and semantics constraints.  
Our evaluators from the internal data team reviewed and annotated all the generated texts with methods PWWS, TextFooler and the translated texts of Checklist dataset, and we finally get three natural test sets, PWWS\textsubscript{nat}, TextFooler\textsubscript{nat} and Checklist\textsubscript{nat}.
\end{itemize}
Besides, we employ PWWS and TextFooler on LCQMC\textsubscript{train} to generate artificial adversarial examples, which are incorporated with original LCQMC\textsubscript{train} as training data (Row \textit{LCQMC+PWWS} and \textit{LCQMC+FOOLER} in Tab.~\ref{tab:adv-results}).The detailed data statistics are shown in Tab.~\ref{tab:adv-statistics-results}. AT details are in Appendix~\ref{appendix: adversarial details}.

\noindent \textbf{Evaluation with artificial and natural adversarial examples.} We fine-tune RoBERTa\textsubscript{l} on LCQMC and the artificial adversarial examples generated by PWWS and TextFooler, and evaluate on the adversarial test sets. The results are shown in Tab.~\ref{tab:adv-results}. Row \textit{LCQMC} shows that only training with LCQMC\textsubscript{train} shows a low performance on \textit{PWWS} and \textit{TextFooler} (we provide a detailed analysis in Appendix~\ref{appendix: black attack}), and the performances on \textit{PWWS} and \textit{TextFooler} are significantly higher on \textit{PWWS\textsubscript{nat}} and \textit{PWWS\textsubscript{nat}}. However, if we incorporate LCQMC\textsubscript{train} with the examples generated by PWWS and TextFooler, the model's performances on \textit{PWWS} and \textit{TextFooler} increase greatly (both methods achieve an great improvement of more than 16\%)
, but the effects on natural examples \textit{PWWS\textsubscript{nat}} and \textit{TextFooler\textsubscript{nat}} are not significant (-5.8\% \textasciitilde2.3\%). On the other 2 natural test sets, Checklist\textsubscript{nat} and DuQM, the effects of 2 adversarial methods are also not obvious (-2.4\% \textasciitilde2.3\%).

In conclusion, the common artificial AT methods are not so effective on the natural datasets. As a corpus consisting linguistically perturbed natural questions, DuQM is beneficial to a robustness evaluation to help us mitigate models' undesirable performance in real-world applications.

\section{Conclusion}\label{sec-conclusion}
In this work, we create a Chinese dataset namely \textbf{DuQM} which contains linguistically perturbed natural questions for evaluating the robustness of QM models. DuQM is designed to be fine-grained, diverse and natural. Specifically, DuQM has 3 categories and 13 subcategories with 32 linguistic perturbation. We conduct extensive experiments with DuQM and the results demonstrate that DuQM has 3 characteristics: 1) DuQM is challenging and has more discrimination ability; 2) 
The fine-grained design of DuQM helps to diagnose the strengths and weakness of models, and enables us to evaluate the models in a diverse; 3)
The effect of artificial adversarial examples does not work on the natural texts of DuQM.

\section*{Ethical Considerations}\label{ethical considerations}
This work presents DuQM, a diverse and natural dataset for the research community to evaluate the robustness of QM models. 
Data in DuQM are collected from a commercial search engine (we are legally authorized by this company), the details are presented in Sec.~\ref{construction overview}. 
Since DuQM do not have any user information, there is no privacy concerns. 
In addition, to ensure that the DuQM is free potential biased and toxic content, we desensitize all the instances in it. 
Regarding to the issue of labor compensation, all the annotators and evaluators are employees from our internal data team and are fairly compensated.


\end{CJK*}

\bibliography{acl}
\bibliographystyle{acl_natbib}

\begin{CJK*}{UTF8}{gbsn}
\appendix
\clearpage
\section{Construction Details}\label{construction details}
Sec.~\ref{construction overview} provides an overview of construction process\footnote{We use \textit{Lexical Analysis of Chinese} (LAC) to do POS tagging, word importance analysis, and NER: https://github.com/baidu/lac. We use a dependency parsing tool: https://github.com/baidu/DDParser} of DuQM. However, DuQM is a diverse dataset with 3 categories and 13 subcategories. And they are constructed with different adversarial methods. Details about our construction approaches to different categories are described in this section.

\noindent\textbf{Lexical Features.}
For each source question, we select the word with specific POS tagss or entity type and high word importance score as \textit{target word}, and perturb the source questions with some other words we collect from following 4 sources:
\begin{itemize}[leftmargin=*,noitemsep]
\item Elasticsearch\footnote{https://github.com/elastic/elasticsearch}: to collect words which have high character overlap with \textit{target words};
\item Faiss\footnote{https://github.com/facebookresearch/faiss}: to collect words which are semantically similar to \textit{target words};
\item Bigcilin\footnote{http://www.bigcilin.com/browser/}: to collect synonym of \textit{target words};
\item Baidu Hanyu\footnote{https://hanyu.baidu.com/}: to collect antonym and synonym of \textit{target words};
\item XLM-RoBERTa\citep{conneau2019unsupervised}: to insert additional words to source sentences\footnote{We add an additional \textit{$\left\{mask\right\} $}
before target word, and use pre-trained language model to predict it. The prediction result of {$\left\{mask\right\} $} is the word inserted to the source sentence.};
\item Vocabulary lists\footnote{Vocabulary lists refer to some word lists containing specific words, such as negation word list and temporal word list.}: to insert some specific words, such as negation word and temporal word.
\end{itemize}

\noindent
\textbf{Syntactic Features.}
For \textit{Symmetry} and \textit{Asymmetry}, we retrieve the source questions in the search log and the returned questions whose edit distance to source question is equal to 4 are selected as candidate questions. Then we compare the dependency structures of the source question and candidate questions. Only the question pairs which contain symmetric or asymmetric relations (which swap the order of two symmetric / asymmetric words) are retained. To generate examples for \textit{Negative Asymmetry}, we select some pairs from \textit{Asymmetry} and negate one side of the pairs. The asymmetric syntactic structure of two sentences and one-sided negation resolves to a positive meaning. For \textit{Voice}, we add "被\textit{bei}" word to source questions to conduct a change of voice. 

\noindent
\textbf{Pragmatic Features.}

\noindent
\textit{Misspelling.}
With the help of Chinese heteronym lists\footnote{https://github.com/FreeFlyXiaoMa/pycorrector/blob/master
/pycorrector/data/same\_stroke.txt}, we obtain a set of common typos and substitute the correct-spelling words with typos. To ensure the correctness, the perturbation should satisfy two constraints:
\begin{itemize}[leftmargin=*,noitemsep]
\item [1)]The typos should be commonly used Chinese characters;
\item [2)]Only one character in the source sentence is replaced with its typo.

\end{itemize}

\noindent
\textit{Discourse Particle.}
We construct this category in 2 ways:
\begin{itemize}[leftmargin=*,noitemsep]
\item [1)]We replace or add some question words, auxiliary words or punctuation marks to generate \textit{Simple Discourse Particle} examples (\textit{Discourse Particle (Simple)} in Tab.~\ref{tab:statistics-exp});
\item [2)]For \textit{Complex Discourse Particle} examples (\textit{Discourse Particle (Complex)} in Tab.~\ref{tab:statistics-exp}), we select some question pairs from a Frequently-Asked-Questions (FAQ) log, especially pairs with big differences in sentence length. Then the pairs are manually annotated and we retained the examples labeled with \textit{Y}.
\end{itemize}

\noindent
With above approaches, we perturb the source questions and obtain a large set of question pairs. Then the generated question pairs are reviewed naturalness and annotated manually.

\begin{figure}[htpb]
  \centering
  \includegraphics[width=0.5\textwidth]{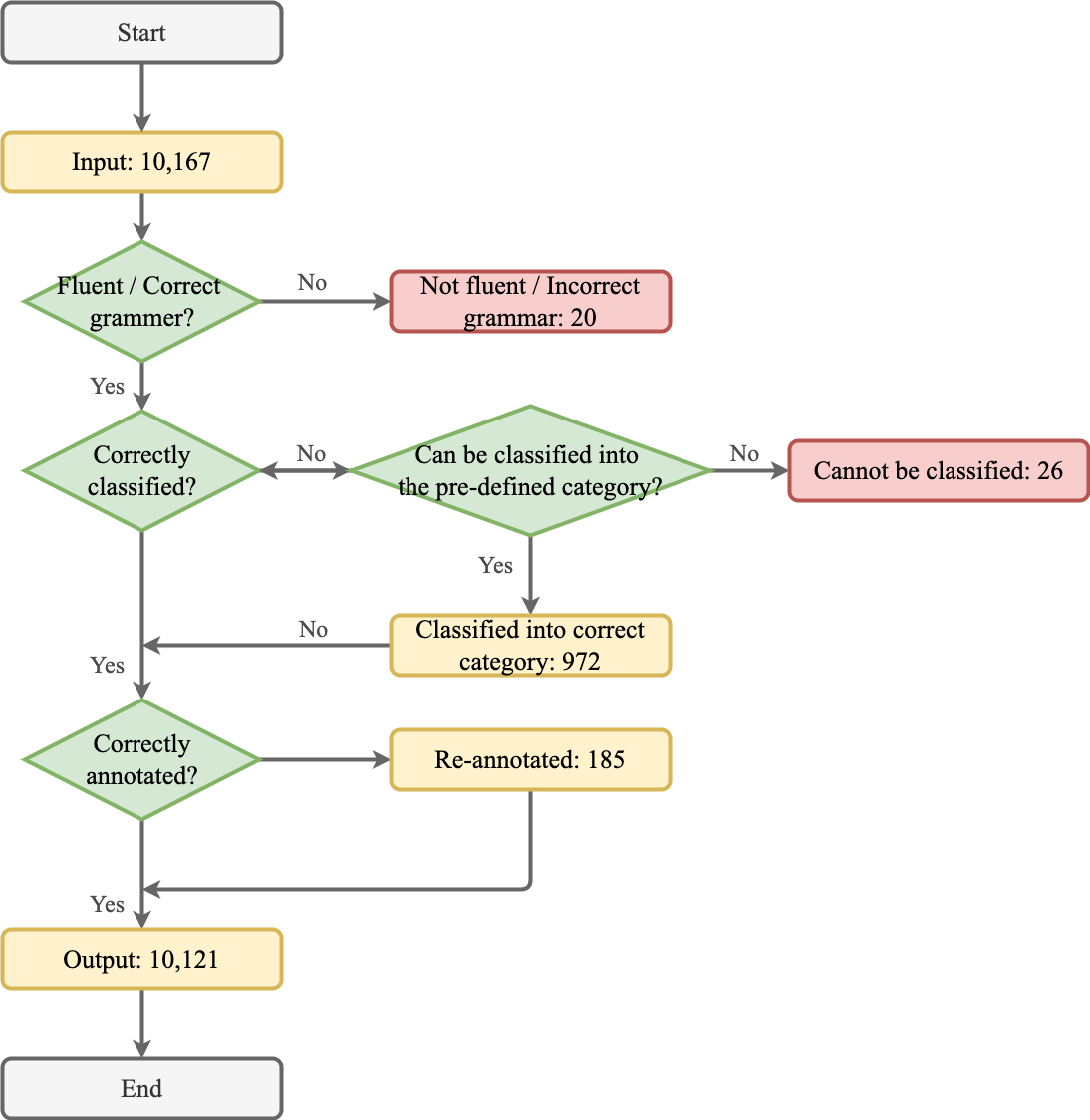}
  \caption{Overall annotation process.} \label{fig:expert_eval}
\end{figure}
\section{Annotation Process}\label{appendix: expert eval}
Fig.~\ref{fig:expert_eval} illustrates the overall annotation process. Only 0.002\% (20/10,167) generated examples are not fluent or not grammatically correct, only 0.1\% (972/10,121) generated examples are re-classified, and only 0.02\% (195/10,121) generated examples are re-annotated manually. 

\begin{table*}
\begin{spacing}{1.4}
\small
\centering
\newcommand{\tabincell}[2]{\begin{tabular}{@{}#1@{}}#2\end{tabular}}
\begin{tabular}{c|c|c|c|c|c|c|c}
\toprule[0.7pt]
Models&L & H & A &\# of Parameters &Masking &LM Task  & Corpus \\
\midrule[0.7pt]
BERT\textsubscript{b} & 12 &768 &12&110M&T&MLM&Wikipedia\\

ERNIE\textsubscript{b} &12 &768 &12&110M&T/E/Ph&MLM&Wikipedia+Baike+Tieba, etc.\\
RoBERTa\textsubscript{b}&12&768&12&110M&MLM&-&EXT\footnote{EXT=Wikipedia+Baike+News+QA corpus, etc.}\\
MacBERT\textsubscript{b}&12&768&12&110M&Mac&SOP&EXT\\

RoBERTa\textsubscript{l}&24&1024&16&340M&MLM&-&EXT\footnote{EXT=Wikipedia+Baike+News+QA corpus, etc.}\\

MacBERT\textsubscript{l}&24&1024&16&340M&Mac&SOP&EXT\\
\bottomrule[0.7pt]
\end{tabular}
\caption{The hyper-parameters of public pre-trained language models we use(L: number of layers, H: the hidden size, A: the number of self-attention heads, T: Token, E: Entity, Ph: Phrase, WWM: Whole Word Masking, NM: N-gram Masking, MLM: Masked LM, Mac: MLM as correction).}\label{tab:pretrain_model_description}
\end{spacing}
\end{table*}




\section{Supplementary Experiments}\label{appendix:experiments}
\subsection{Additional Experimental Setting}
\subsubsection{Training Details}\label{appendix:train details}
In the fine-tuning stage, we insert a $[SEP]$ between the question pairs. The pooled output is passed to a classifier. 
We use different different learning rates and epochs for different pre-trained. Specifically, for large models, the learning rate is 5e-6 and the number of epochs is 3. For base models, the learning rate is 2e-5, and we set the number of epochs as 2. The batch size is set as 64 and the maximal length of question pair is 64. 
We use early stopping to select the best checkpoint. Each model is fine-tuned 3 times on LCQMC\textsubscript{
train} and we choose the model with the best performance on LCQMC\textsubscript{dev} to report test results. 

\subsubsection{Datasets Details} 
Tab.~\ref{tab:statistics-LCQMC} gives a detailed description of LCQMC Corpus. 


\begin{table}
\begin{spacing}{1.4}
\small
\centering
\newcommand{\tabincell}[2]{\begin{tabular}{@{}#1@{}}#2\end{tabular}}
\begin{tabular}{c|c|c|c|c}
\toprule[0.7pt]
Corpus&Train & Dev & Test & Fine-grained  \\
\midrule[0.7pt]
LCQMC & 238,766 &8,802 &12,500&No\\
\bottomrule[0.7pt]
\end{tabular}
\caption{Data statistics of LCQMC.}\label{tab:statistics-LCQMC}
\end{spacing}
\end{table}
\begin{table}
\begin{spacing}{1.4}
\small
\centering
\newcommand{\tabincell}[2]{\begin{tabular}{@{}#1@{}}#2\end{tabular}}
\begin{tabular}{l|cc}
\toprule[0.7pt]
 \textbf{Data} & BERT & RoBERTa \\
\midrule[0.7pt]

PWWS & \underline{41.5} & \underline{41.9} \\
PWWS\textsubscript{nat} & 23.0\textcolor{textred}{\textsubscript{-18.5}} & 18.5\textcolor{textred}{\textsubscript{-23.4}} \\
\midrule[0.4pt]
TEXTFOOLER & \textbf{46.6} & \textbf{42.9}  \\
TEXTFOOLER\textsubscript{nat} & 14.6\textcolor{textred}{\textsubscript{-32.0}} & 12.2\textcolor{textred}{\textsubscript{-30.7}}  \\
\midrule[0.4pt]
DuQM & 33.4 & 26.2  \\

\bottomrule[0.7pt]

\end{tabular}
\caption{Attack success rate(\%) on different test data.}\label{tab:unnatural-results}
\end{spacing}
\end{table}
\subsection{Adversarial Training Details}\label{appendix: adversarial details} 
Tab.~\ref{tab:adv-statistics-results} gives a detailed statistics of adversarial examples generated with TextFooler, PAWS. To generate training samples, we select a set of LCQMC training questions and apply the methods PWWS and TextFooler on them. The labels are same as original samples. To generate test samples and ensure a robust evaluation, we utilize 4 datasets, PWWS\textsubscript{nat}, TextFooler\textsubscript{nat}, Checklist\textsubscript{nat}\footnote{Before annotating, we translate original Checklist dataset into Chinese using a translation tool} and DuQM, which are natural adversarial examples. We conduct an experiment about adversarial training by feeding the models both the original data and the adversarial examples, and observe whether the original models become more robust. We use pre-trained model RoBERTa\textsubscript{l} (described in Tab.~\ref{tab:pretrain_model_description}) for fine-tuning and the fine-tuning details are described in Sec.~\ref{sec:experimental setup}. 
\subsection{Results of Attacks}\label{appendix: black attack} 
We give the main results of attacks to BERT\textsubscript{b} and RoBERTa\textsubscript{l} in Tab.~\ref{tab:unnatural-results}. The results show that the un-natural attacks (on artificial adversarial samples, i.e. PWWS and TextFooler in Tab.~\ref{tab:unnatural-results}) have higher success rate than DuQM.
However, if we select the natural examples from the artificial adversarial samples (PWWS\textsubscript{nat} and TextFooler\textsubscript{nat} in Tab.~\ref{tab:unnatural-results}), the attack success rate of PWWS and TextFooler is significantly decreasing by at least 18.5\% on BERT\textsubscript{b} and 30.7\% on RoBERTa\textsubscript{l} respectively. DuQM, in which all the samples are natural and grammarly correct, gets the best performance when black-box attacking (compare to PWWS\textsubscript{nat} and TextFooler\textsubscript{nat} in Tab.~\ref{tab:unnatural-results}). In summary, the artificial adversarial examples training is not effective on natural texts, such as DuQM. It is reasonable that we should pay more attention to the naturalness when generating the adversarial examples.

\end{CJK*}

\end{document}